\documentclass[graybox]{svmult}

\usepackage{mathptmx}       % selects Times Roman as basic font
\usepackage{helvet}         % selects Helvetica as sans-serif font
\usepackage{courier}        % selects Courier as typewriter font
\usepackage{type1cm}        % activate if the above 3 fonts are
\usepackage{float} % added for mini-page split in Train On target
\usepackage{makeidx}         % allows index generation
\usepackage{graphicx}        % standard LaTeX graphics tool
\usepackage{multicol}        % used for the two-column index
\usepackage{multirow}
\usepackage{makecell}
\usepackage{tikz,graphics,color,float,epsf,caption,subcaption}
\usepackage[bottom]{footmisc}% places footnotes at page bottom
\makeindex             % used for the subject index
\begin{document}

\title*{Language Portability Strategies for Open- domain Dialogue with Pre-trained Language Models from High to Low Resource Languages}
\titlerunning{Language portability strategies for open-domain dialogue with PLMs}
\author{Ahmed Njifenjou, Virgile Sucal,
Bassam Jabaian and Fabrice Lefèvre}
\authorrunning{A. Njifenjou, V. Sucal, B. Jabaian, F. Lefèvre}
\institute{Ahmed Njifenjou, Virgile Sucal, Bassam Jabaian, Fabrice Lefèvre \at LIA/CERI, Avignon Université, France 
\and
{ \{ahmed-ndouop.njifenjou, virgile.sucal, bassam.jabaian, fabrice.lefevre\}@univ-avignon.fr}}
\maketitle

\abstract*{In this paper we propose a study of linguistic portability strategies of large pre-trained language models (PLMs) used for open-domain dialogue systems in a high-resource language for this task. In particular the target low-resource language ($L_T$) will be simulated with French, as it lacks of task-specific resources  and allows our human evaluation, when the source language ($L_S$) is English. For obvious reasons, recent works using such models for open-domain dialogue are mostly developed in English. Yet building specific PLMs for each possible target language supposes collecting new datasets and is costly.  For this reason, trying to leverage all existing resources (PLMs and data) in both $L_S$ and $L_T$, we wish to assess the performance achievable in $L_T$ with different approaches. The first two approaches evaluate the usage of Neural Machine Translation (NMT) at different levels: TrainOnTarget where a $L_S$ dataset is translated before fine-tuning in $L_T$ and TestOnSource where a $L_S$ model is coupled with NMT modules during inference. 
Then, the advent of BLOOM~\cite{bigscience_workshop_2022},
the world first  \textbf{open-access} multilingual large PLM, allow researchers to develop new approaches aiming to leverage not only the model's full accessibility but also its multilingualism and translation abilities. In this context the task is learned in $L_S$ first and adapted to $L_T$ using the MAD-X Adapter architecture~\cite{Pfeiffer2020}. In the two sets of experiments models are evaluated in spoken dialogue conditions with human and the strategies can be compared in terms of perceived interaction quality.}

\abstract{In this paper we propose a study of linguistic portability strategies of large pre-trained language models (PLMs) used for open-domain dialogue systems in a high-resource language for this task. In particular the target low-resource language ($L_T$) will be simulated with French, as it lacks of task-specific resources  and allows our human evaluation, when the source language ($L_S$) is English. For obvious reasons, recent works using such models for open-domain dialogue are mostly developed in English. Yet building specific PLMs for each possible target language supposes collecting new datasets and is costly.  For this reason, trying to leverage all existing resources (PLMs and data) in both $L_S$ and $L_T$, we wish to assess the performance achievable in $L_T$ with different approaches. The first two approaches evaluate the usage of Neural Machine Translation (NMT) at different levels: TrainOnTarget where a $L_S$ dataset is translated before fine-tuning in $L_T$ and TestOnSource where a $L_S$ model is coupled with NMT modules during inference. 
Then, the advent of BLOOM~\cite{bigscience_workshop_2022}, 
the world first  \textbf{open-access} multilingual large PLM, allow researchers to develop new approaches aiming to leverage not only the model's full accessibility but also its multilingualism and translation abilities. In this context the task is learned in $L_S$ first and adapted to $L_T$ using the MAD-X Adapter architecture~\cite{Pfeiffer2020}. In the two sets of experiments models are evaluated in spoken dialogue conditions with human and the strategies can be compared in terms of perceived interaction quality.}

    \section{Introduction}
\label{sec:1.0}
Since the breakthrough of transformers~\cite{Vaswani2017}, the field of Natural Language Processing (NLP) has been fuelled with many variants of PLMs. Auto-regressive transformers, those using the decoder part of the transformers, like GPT~\cite{Radford2018ImprovingLU}, BART~\cite{Lewis2019} and derivatives helped to improve the state-of-the art of several generative tasks. Among these open-domain dialogue systems, aka chatbots, which as stated in~\cite{walker2021} should develop some human abilities like empathy, personality and entertainment to socially engage with users. For this extent, specific crowd-sourced datasets have been created (such as PersonaChat~\cite{Zhang2018d},  Empathetic Dialogues~\cite{rashkin2019towards}, Blended Skill Talk~\cite{Smith2020} etc.) on which PLMs are fine-tuned. Unfortunately, most of these resources are in English. Even French which is generally not considered as a low-resource language lacks of specific datasets for this task.

In this work we study portability strategies of chitchat models and datasets from a source language ($L_S$, English here) to a target language ($L_T$, French here), with the broad target of later application to truly low-resource language. Making the most of resources available in both $L_S$ and $L_T$ - NMT tools, datasets and PLMs models - we set up and conducted a human evaluation of different systems obtained by combining those resources in three different approaches, described hereafter, varying their usage of the resources: training on target, testing on source and training on source then adapting to target. All the models are compared to an overall reference model in $L_S$; BlenderBot 1.0~\cite{roller2020recipes} a much advanced system in terms of architecture and datasets is retained. 

\section{Related Work}
\label{sec:2.0}

In previous team works on domain and language portability of spoken dialogue systems~\cite{Jabaian2013, Lefevre2010}, we assessed language portability of Spoken Language Understanding module of a goal-oriented dialogue system. A Statistical Machine Translation (SMT) module, state-of-the-art back then, was used in two approaches efficient and low-cost for language portability. Here we assess these strategies with state-of-the-art NMT modules on transformers-based open domain dialogue systems.

Coming as one of the rare works that tackled the issues of monolingual resources development for open domain chatbot,~\cite{Zhaojiang2020} is the closest work to our current study - to the best of our knowledge. Indeed they also worked with PersonaChat and provided its translated versions in French and five additional languages. Furthermore they assessed multilingual and cross-lingual approaches with a specific target on multilingual models. They also evaluated a model using translation before and after a $L_S$ model inference, however with a model with a different architecture from the others. But automatic translation of PersonaChat were revised by native $L_T$ speakers that are fluent in $L_S$, which is costly and cannot be afforded for some languages where bilingual speakers can be scarce. Besides, their cross-lingual model had poor results compared to others. Also our strategies implement another approach, already used in~\cite{Pfeiffer2020} to perform cross-lingual transfer learning on a bunch of NLP tasks but not including generative tasks. They used a modular adapters framework, baptised MAD-X, combined to a multilingual PLM. The latter is frozen while its blocks are augmented with unfrozen language and task adapters. These adapters are then trained sequentially: first the language adapters which in turn are frozen before the task adapters training. This approach can be less expensive in terms of training cost when adding a new language support. Also with task adapters primarily trained on $L_S$ original data rather than their noisy auto-translations counterparts, it may help improve performance.

For dialogue modeling, we used the same training framework as~\cite{wolf2019} on PersonaChat. Model inputs here are a concatenation of model-assigned personality traits, dialogue history and the ``golden" reply it learns to mimic. In addition to what they did with a GPT~\cite{Radford2018ImprovingLU} model, we tried the same approach in French and with the BLOOM~\cite{bigscience_workshop_2022} model which is an open access and multilingual model.

\section{Portability Strategies Assessed}
\label{sec:3.0}
In this preliminary study of our ongoing work on French and other languages lacking specific data for open-domain dialogue, rather than focusing on intrinsic dialogue performance improvement, we assess how the data and models from $L_S$ can be leveraged to develop shallow conversational models based on transformers in $L_T$.
 
\subsection{TestOnSource and TrainOnTarget} \label{sec:3.1}
Inspired from our previous work on cross-lingual Spoken Language Understanding~\cite{Jabaian2013, Lefevre2010}, these two approaches rely on the usage of NMT modules at different stages. Recent advances in this field provide us with ready to use high quality translation tools for $L_S$ and $L_T$. The \texttt{DeepL} tool claimed to be the best performing\footnote{https://www.deepl.com/en/quality.html} is only accessible through a paying (for more than 500K characters) API, hence we used \texttt{Google Translate} API which is the same API used in \cite{Zhaojiang2020}.

\begin{center}
    \begin{figure}[h]
        \sidecaption
        \includegraphics[scale=.6]{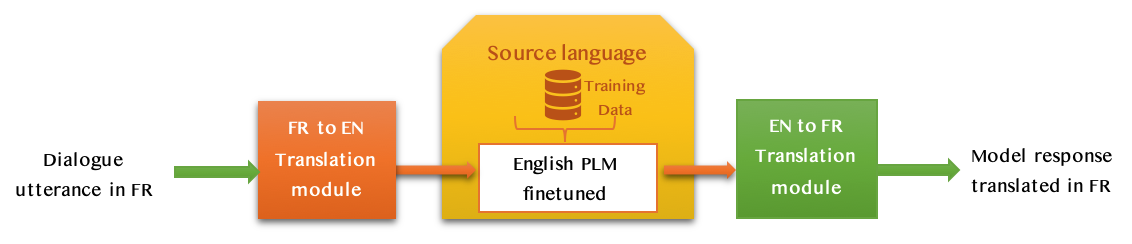}
        \caption{\textbf{Illustration of the TestOnSource approach:} an additional $L_T$ to $L_S$ (orange) and $L_S$ to $L_T$ (green) NMT modules around a $L_S$ model fine-tuned for chitchat}
        \label{fig:3.1}      
    \end{figure}
\end{center}

\subparagraph{Test On Source}
The large amount of resources including models for open-domain dialogue in $L_S$ is a major asset. Consequently, it is interesting to evaluate how well these systems perform on  inputs translated from $L_T$ to $L_S$ at inference. Hence the approach merely consists in using existing datasets, PLM and open-domain dialogue models available in $L_S$ and combine them with NMT system during bot-human conversations in $L_T$ as shown in Fig.~\ref{fig:3.1}. 
  
\subparagraph{Train On Target}
    \begin{minipage}{0.38\textwidth}
         While lacking open-domain dialogue specific datasets, the $L_T$ in this study - French - has at its disposal a bunch of PLMs that can be used as a basis for a dialogue system. TrainOnTarget, illustrated in Fig. \ref{fig:3.2}, consists in fine-tuning adequate $L_T$ PLMs (bottom green) on chitchat task using an automatically translated dataset from $L_S$ (top yellow). Despite being subject to noise injection in data, forward translation can still be effective for low-resource MT in some contexts~\cite{haddow:hal-03479757} and we assume that the language specific abilities learned by the $L_T$ PLMs can help handle the noisy NMT samples. 
    \end{minipage}\hfill \hfill
   \begin{minipage}{0.59\textwidth}
        \begin{figure}[H]
            \includegraphics[scale=.6]{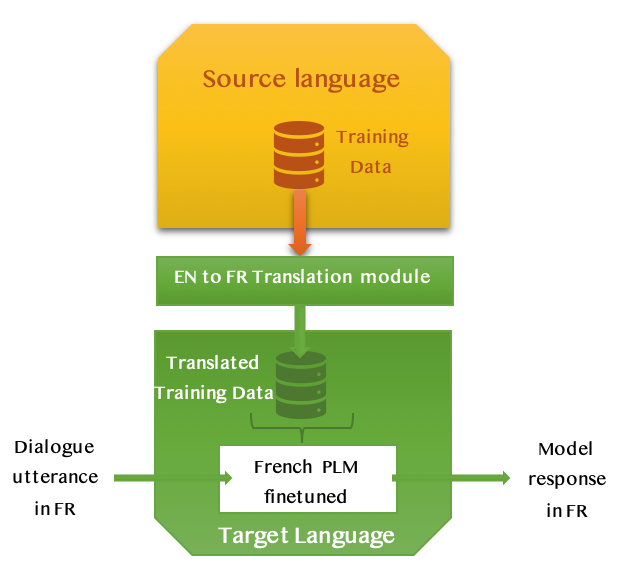}
            \caption{ \textbf{Illustration of the TrainOnTarget approach}}
            \label{fig:3.2}      
         \end{figure}
    \end{minipage} 

\subsection{TrainOnSourceAdaptOnTarget: leverage multilingual PLMs}
The two previous approaches rely heavily on the fact that outside of chitchat task $L_T$ is not under-resource having available NMT tools and PLMs. Hence, the idea of using  multilingual PLMs to not exclude the large majority of lower-resource languages.  

\begin{center}
    \begin{figure}[h]
        \sidecaption
        \includegraphics[scale=.57]{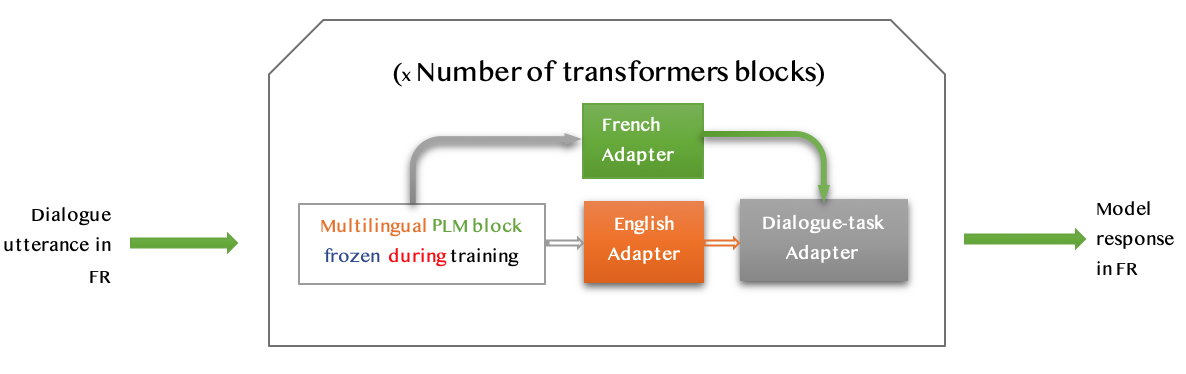}
        \caption{ A transformer block in \textbf{TrainOnSourceAdaptOnTarget} configuration. Empty arrows show the first fine-tuning stage of the task adapters (on $L_S$ data) with $L_S$ language adapters frozen. In the second stage (filled arrows) language adapters (still frozen) are switched from $L_S$ to $L_T$ and the same task adapters (assumed language agnostic~\cite{Pfeiffer2020}) are fine-tuned using few (or translated) $L_T$ data. Prior to these stages, both $L_S$  and $L_T$ language adapters are trained independently on causal LM with transformers block frozen.}
        \label{fig:3.3}      
    \end{figure}
\end{center}

We reproduce the MAD-X architecture~\cite{Pfeiffer2020} for dialogue using \textbf{BLOOM} which has translation abilities over a large set of low-resource languages. This is interesting as it can help translate subset of $L_S$ datasets which will then be used for dialogue-task adapter's few-shot tuning. The latter being firstly finetuned on human generated data in $L_S$ rather than directly on NMT resulting data (as in \ref{sec:3.1}) may also be a good asset. Fig. \ref{fig:3.3} shows the overall workflow of this approach.

\section{Experimental Setup} \label{sec:4.0}
Having the previous portability approaches in mind, we trained and evaluated different PLMs from $L_S$ and $L_T$ and eventually multilingual with the recent advent of \textbf{BLOOM}. The target was first to assess the most efficient approach with the current resources but also to compare to the previous results on similar approaches.

\subsection{PersonaChat Dataset} \label{sec:4.1}
To train and evaluate the different models we chose the PersonaChat~\cite{Zhang2018d} dataset described in Table \ref{tab:4.1}. It consists of a set of dialogues between two humans in English, each being assigned a personality defined by a few sentences. The aim of this dataset is to help build models with consistent persona throughout a conversation. For French-based models, the dataset is automatically translated with the \texttt{Google Translate}  API\footnote{https://translate.google.fr}. 

However, contrary to~\cite{Zhaojiang2020}, no human re-annotation was performed after auto-translation as they reported a BLEU of \textbf{94.19} while assessing the difference between auto-translated and human corrected data in French. Also we didn't re-use their corrected dataset in order to mimic the case of a language where no native speakers fluent in English can be hired to correct translation errors.

\begin{table}
\caption{Statistics of the PersonaChat dataset used for models' finetuning}
\label{tab:4.1}       
\begin{tabular}{p{3.5cm}p{3.5cm}p{2cm}p{2cm}}
\hline\noalign{\smallskip}
\textbf{Language} & \textbf{Split} & \textbf{Dialogues} & \textbf{Utterances}  \\
\noalign{\smallskip}\svhline\noalign{\smallskip}
\multirow{3}{*}{\textbf{English ($L_S)$}} & Train & 14917$^*$ & 109718\\
                                        & Validation & 100 &  773 \\
                                        & Test & 1000 & 7801 \\
\hline
\multirow{3}{*}{\textbf{French ($L_T$)}} & Train & 14917 & 109759\\
                                        & Validation & 100 &  773 \\
                                        & Test & 1000 & 7801 \\

\noalign{\smallskip}\hline\noalign{\smallskip}
\end{tabular}

$^*$ During translation some problematic conversations were dropped, so for $L_S$ we took the same training size as the $L_T$ version to be fair to the corresponding trained models. The original dataset has more than 17K dialogs.
\end{table}

\subsection{Models' Descriptions} \label{subsec:4.2}
We consider open-domain dialogue here as a generative task, hence we focused on leveraging  auto-regressive models available in both $L_S$ and $L_T$.

\runinhead{Monolingual-PLM based models} For \textit{ TrainOnTarget} we fine-tuned the small version of \textbf{GPT-fr}~\cite{simoulin:hal-03265900} (124M parameters), a French version of GPT-2~\cite{radford2019language} as it is close in terms of architecture and size to the GPT-based TransferTransfo~\cite{wolf2019} model  (117M parameters) that will be used for monolingual \textit{TestOnSource}. 

\runinhead{Multilingual-PLM based models} The model used here is \textbf{BLOOM} a brand new large Multilingual PLM accessible to researchers.  We worked on its thinnest version available which has 560M parameters. Multilingual PLMs are the only to allow the third approach, hence we built a $L_S$ and $L_T$ models using MAD-X adapters architecture on BLOOM. The resulting models are reported as \textit{madx-BLOOM} in Table \ref{tab:5.1}. As the model is multilingual, we also built a $L_S$ and $L_T$ models for the first two approaches i-e without using adapters mixing across the languages which are reported as \textit{BLOOM ($L_T$)} and  \textit{BLOOM ($L_S$)} in Table \ref{tab:5.1}.

\subsection{Training Details}
All the models were trained using a double-heads architecture as in ~\cite{wolf2019}: a Causal Language Modeling head and a multi-choice head. The former, had a higher weight on the combined loss as we assumed the dialogue to be mainly a generative task.

The GPT-fr based model, as its $L_S$ counterpart~\cite{wolf2019}, was fine-tuned for one epoch with AdamW optimizer and a linear-decreasing learning rate of 6.25e-5 on PersonaChat translated. Both BLOOM\_fr and BLOOM\_en were trained with that same learning rate for 5 epochs with evaluation performed every quarter of an epoch and the 5 five checkpoints with lowest perplexity kept. Then the best performing checkpoint on the whole test set was retained.

For the models with the adapter mixing architecture, we first trained the $L_S$ and $L_T$ language adapters on Wikipedia\footnote{The following pre-processed subsets of Wikipedia available on HuggingFace were used: 20220301.fr ($L_T$) and 20220301.en ($L_S)$} with a total batch size of 80 and a relatively high initial learning rate of 1e-4 following ~\cite{Pfeiffer2020} for 379K and 427k steps respectively (one week on five V100). As the validation perplexity was still decreasing, we used the last checkpoints as language adapters to train the task adapters. First they were trained on PersonaChat in $L_S$, then the language adapters were switched from $L_S$ to $L_T$ before the fine-tuning of the unchanged task adapters on the translated PersonaChat for the $L_T$ model.

\section{Evaluation}
\label{sec:5.0}
As assumed in~\cite{zhao2017} dialogue has a one-to-many structure which makes automatic metrics based on word overlaps often not correlate with human evaluations as they can dismiss good dialogue utterances that are different from the ground-truth~\cite{Mehri2020, Gupta2019}. 
Hence human evaluation remains the most reliable but some automatic metrics like perplexity as shown in~\cite{Adiwardana2020} can sometimes be somehow correlated to human judgements, so we add the information. 

\subsection{Automatic Metrics} \label{subsec:5.1}
We evaluated each of the models using the test set of PersonaChat in the relevant language. We computed perplexity for our models as it is available for all other state-of-the-art models. In addition Hits@1/3 was computed when possible (models trained, with a multi-choice head). This metric represents the accuracy of ranking the next gold utterance first in a set with two distractors.

We used greedy decoding to compute BLEU score~\cite{papineni-etal-2002-bleu} for comparison with other models when available, but also to show how it is at odd with actual dialogue model performance while perplexity is indicative. All the results of automatic metrics are reported in Table~\ref{tab:5.1}. 

\begin{table}
\caption{Automatic evaluations of the different models grouped by strategy}
\label{tab:5.1}     
\begin{tabular}{p{3cm}p{3cm}p{2cm}p{1.5cm}p{1.5cm}}
\hline\noalign{\smallskip}
\textbf{Strategy} & \textbf{Models}$^*$ & \textbf{Perplexity }& \textbf{Hits@1}$^{**}$ & \textbf{BLEU} \\
\noalign{\smallskip}\svhline\noalign{\smallskip}
\multirow{2}{*}{\textbf{Train On Target}} & GPT-fr & \textbf{10,82} & 0,88 & N/A \\ 
                                        & BLOOM & 16,05 & \textbf{0,95} & 0,23 \\
\hline
\multirow{2}{*}{\textbf{Test On Source}} & \textit{GPT}~\cite{wolf2019}  & 18,49 & 0,84 & N/A \\
                                       & BLOOM  & \textbf{13,01} & \textbf{0,94} & 0,22\\
\hline                          
\multirow{4}{*}{\textbf{CrossLingual Training}} &  \textit{XNLG}$^{***}$ ($L_S$) & 54,74 & N/A & \textbf{2,25} \\
                                       &  madx-BLOOM ($L_S$)  & \textbf{24,07} & 0,82 & 0,13\\

                                        \cline{2-5}
                                       &  \textit{XNLG} ($L_T$) & 640.33 & N/A & 0,09 \\
                                       &  madx-BLOOM ($L_T$) & \textbf{28,64} & 0,81 & \textbf{0,15}\\

\noalign{\smallskip}\hline\noalign{\smallskip}
\end{tabular}

$^*$ Models in italics and metrics associated are state-of-the-art

$^{**}$ Available for models trained with double head

$^{***}$ Both XNLG model metrics are from~\cite{Zhaojiang2020}
\end{table}

In~\cite{Zhaojiang2020}, they reported extremely bad results on automatic metrics for models trained in cross-lingual framework. The MAD-X architecture on BLOOM improves these metrics with a significant gain especially on $L_T$ models: from 640 to 28 on perplexity while we actually evaluate on the whole PersonaChat test set compared to a subset in their case. A lower perplexity can guarantee some generation capabilities, however the generated ouputs may often be out of context in a dialogue framework yielding poor results in human evaluation.

\subsection{Human Evaluation} \label{subsec:5.2}
We performed  dialogue collection using the RASA-X~\cite{DBLP:journals/corr/abs-1712-05181} platform. It was done in two phases: in the first phase we deployed \texttt{GPT-fr}, \texttt{TransferTransfo(GPT)} and \texttt{BlenderBot 1} and in the second all the four models based on \texttt{BLOOM}. We collected 140 conversations\footnote{For research purposes all collected data can be requested for by an e-mail to the first author.} and evaluated them on three criteria selected based on those in~\cite{Mehri2020, Ji2022-kr, roller2020recipes}: coherence, engaging-ness and humanness. Detailed description of the dialogue collection and annotation process with an analysis of inter-annotators agreement are reported in Appendix. 

\subsubsection{Overall Ratings}
Fig.~\ref{figs:ratings} reports the overall ratings of the conversations for each model and by assessed quality\footnote{ All are French ($L_T$) conversations generated and evaluated by French speakers even when the original conversational model is in English ($L_T$) (BlenderBot, GPT\_EN, BLOOM\_en and madxBLOOM\_en in Fig. \ref{figs:ratings}).}. For each conversation, we averaged 3 evaluators' ratings per criterion. 

Without a suprise \texttt{BlenderBot 1} sets the reference. It's a bigger model ($\sim$ 2.7B parameters distilled into 400M) and it is trained on larger and varied datasets (Blended Skill Talk, Empathetic Dialogues, Wizard of Wikipedia~\cite{Dinan2018} and also PersonaChat) with complex learning objectives. There is neither a comparable model in $L_T$, nor equivalent datasets. Hence this study of portability strategies focused on the other smaller models as an entry before generalizing to bigger and more complex models.

\begin{figure}
    \begin{subfigure}{0.33\linewidth}
        \centering
        \includegraphics[width=1\linewidth]{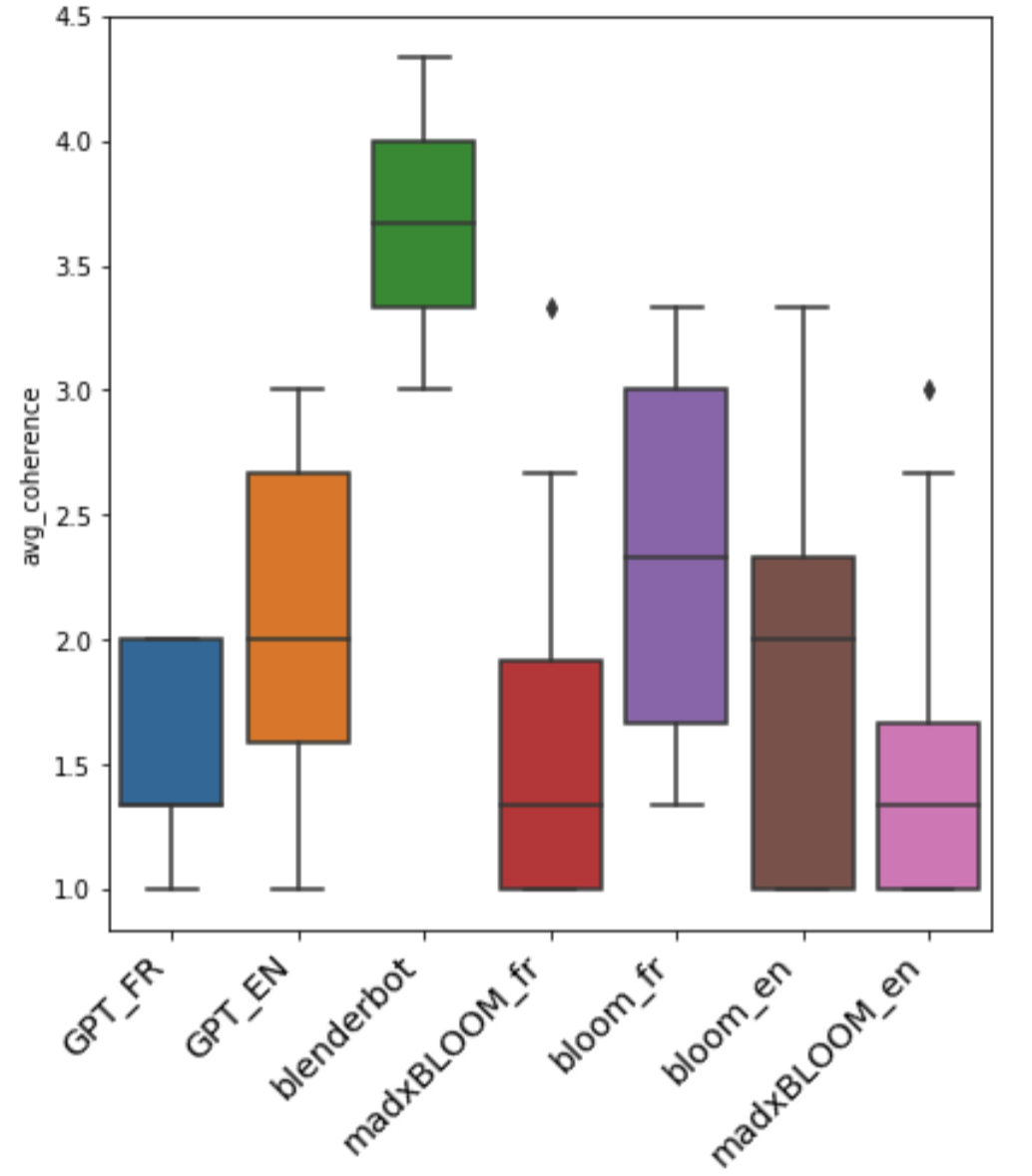}
         \caption{Coherence} 
         \label{figs:coherence}
    \end{subfigure}
    \begin{subfigure}{0.33\linewidth}
        \centering
        \includegraphics[width=1\linewidth]{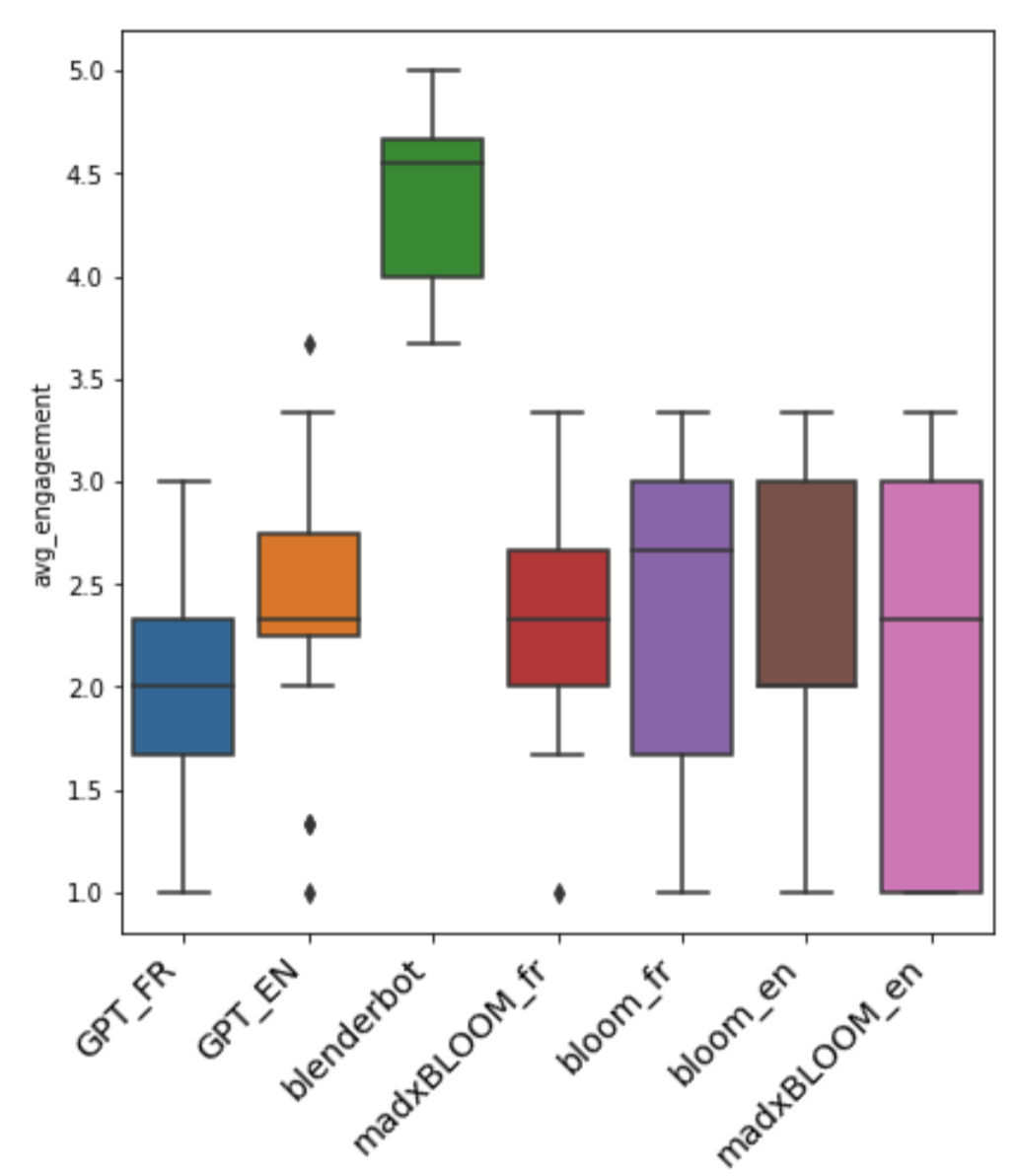}
         \caption{Engagingness}
         \label{figs:engagement}
    \end{subfigure}
    \begin{subfigure}{0.33\linewidth}
        \centering
        \includegraphics[width=1\linewidth]{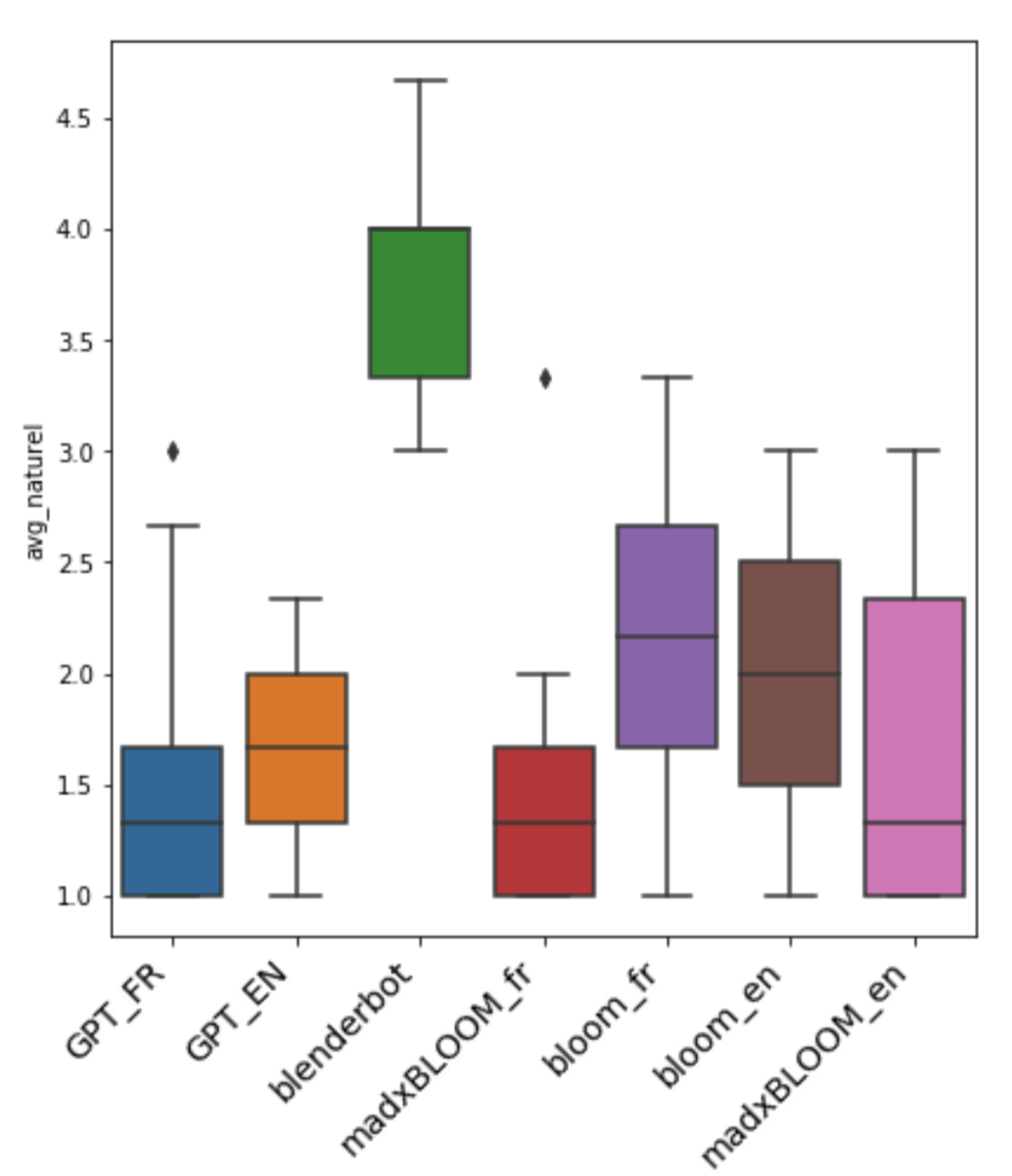}
        \caption{Humanness} 
        \label{figs:Humanness}
    \end{subfigure}
    \caption{Average overall ratings for the seven models deployed. In green is reference BlenderBot 1}
    \label{figs:ratings}
\end{figure}

Out of the six remaining models (with \texttt{GPT\_EN} being from the state-of-the-art), \texttt{BLOOM\_fr} emerges as the best in all three assessed categories in average: \textbf{0.25} on coherence, \textbf{0.07} on engagement and \textbf{0.3} on humanness from its closest runner-up \texttt{GPT\_EN}. The latter has close ratings in average with \texttt{BLOOM\_en}. The former has a marginal advantage on coherence (0.04) and engagingness (0.06) while the latter is slighly better on humanness (0.3). Then we have the last group composed by \texttt{GPT\_FR}, \texttt{madxBLOOM\_fr}, \texttt{madxBLOOM\_en} in which  the median average rating in all categories is below 1.5 meaning nearly half of the conversation generated with these models were rated to the lowest. 

\begin{table}
\caption{Average number of utterances per model}
\label{tab:5.2.2}     
 \resizebox{\textwidth}{!}{
\begin{tabular}{p{1.5cm}|c|c|c|c|c|c|c}
\hline\noalign{\smallskip}
\textbf{Model} &  \textbf{Blender} & \textbf{GPT\_FR} & \textbf{GPT} & \textbf{xBLOOM\_fr}  & \textbf{xBLOOM\_en} & \textbf{BLOOM\_fr}  & \textbf{BLOOM\_en} \\
\noalign{\smallskip}\svhline\noalign{\smallskip}
\textbf{\#utterances}  & 35.6 & 15.4 &  \textbf{24.8} & 12.9  & 20.6 & \textbf{ 24.8} &  \textbf{36.3}   \\
\noalign{\smallskip}\hline\noalign{\smallskip}
\end{tabular}
}
Where Blender stands for BlenderBot, xBLOOM for model with MAD-X architecture applied to bloom and GPT is for the English model based on GPT from \cite{wolf2019}.
\end{table}

 Table \ref{tab:5.2.2} show statistics on collected  dialogues average number of utterances. They give another view of the models performances. Once again the BlenderBot sets the reference.  We can notice that, the lowest rated models are also the one with the lowest average number of utterances with madxBLOOM\_fr having the worst with 12.9 utterances in average. Not only it explains the relatively low engagingness ratings of these models but also the coherence score. Indeed, early ending conversations with respect to our guidelines, indicate that some hallucinations and erratic behaviours like repetitiveness and dullness have been observed by the users (examples of theses behaviour are provided in Table \ref{tab:5.2.1}). The best models emerge with an average number of utterances per conversation higher ranging from \textbf{24.8} to \textbf{36.3}.

\begin{table}
\caption{Examples of hallucinations and erratic behaviours in a dialogue }
\label{tab:5.2.1}       

\begin{tabular}{p{4cm}p{3.7cm} p{3.7cm}}
\noalign{\smallskip}\svhline\noalign{\smallskip}
Bot Utterance  &  \textit{Translated utterance} &  User utterance \\
\noalign{\smallskip}\svhline\noalign{\smallskip}
\underline {\texttt{\textbf{madxBLOOM\_fr}}}  & & \\
& \textit{\textbf{Hello, I may need your help}}  & Salut, j'aurais besoin de ton aide \\
bien je passe un peu de temps à enseigner à l'université de ma ville. tu? & ~ \textit{\textcolor{red}{Well I spend sometimes teaching at my city's University. And you?}} &   \\

& \textit{\textbf{You teach Maths I hope}} & Tu enseignes des maths j'espère  \\
abétique à l'école. J'aimerais que vous êtes d'avantage avec lui & \textcolor{red}{\textit{"Abétique" at school. I want you to be even more with him}}  & \\

& \textit{\textbf{Well, I have some difficulties with my homeworks}} & et bien j'ai des difficultés avec mes devoirs  \\

Je porte souvent de promener dans la musique et à jouer de la musique en France & \textit{ \textcolor{red}{I wear often wandering in the music and playing some music in France} } &   \\

\noalign{\smallskip}\svhline\noalign{\smallskip}
\underline {\texttt{\textbf{GPT\_fr}}}  & & \\

Je suis un graphiste et un écrivain & \textit{I am a \textcolor{teal}{graphist} and a \textcolor{blue}{writer}}  &\\

& \textit{\textbf{Graphist and writer!}} &  Graphiste et écrivain ! \\

c'est génial. J'aime lire & \textit{it's \textcolor{teal}{awesome}. I like \textcolor{blue}{reading}} &   \\

 & \textit{\textbf{What kind of book you say[read]}}  & quel genre de livre tu dis[lis]$^*$ \\
 
Je suis un écrivain créatif & \textit{I'm a \textcolor{teal}{creative} \textcolor{blue}{writer}} &\\

& \textit {\textbf{Do you write?}} & Est ce que tu écris? \\

Je suis un artiste & \textit{I'm an \textcolor{teal}{artist}} &\\

& \textit {\textbf{Me too}} & Moi aussi \\

Je suis un écrivain & \textit{I'm a \textcolor{blue}{writer}} &\\

\noalign{\smallskip}\svhline\noalign{\smallskip}
\underline {\texttt{\textbf{madxBLOOM\_en}}}  & & \\
& \textit{\textbf{Hi! I like climbing}} & Salut! J'aime l'escalade \\
Moi je n'aime pas courir & \textit{I don't like running}  & \\
& \textit{\textbf{What sport do you practice?}} & Quel sport pratiques-tu?,  \\
J'aime aller à la gym et me détendre & \textit{I like to go to the gym and relax}  & \\

& \textit{\textbf{Are you a gymnast?}} & Tu es gymnaste ? \\
Tu fais quoi comme boulot & \textit{ \textcolor{red}{What do you do as a job?} } &   \\
& \textit{\textbf{Answer my question}} & Réponds à ma question \\
j'ai travaillé dans un bar, mais je ne travaille pas & \textit{ \textcolor{red}{I worked in a bar but I don't work} } &   \\
& \textit{\textbf{Why don't you work?}} & Pourquoi tu ne travaille pas? \\
C'est ce que je fais dans la vie & \textit{ \textcolor{red}{That's what I do for a living} } &   \\
& \textit{\textbf{Are you tired of life?}} & Tu es fatigué de la vie ? \\
oui c'est un super metier & \textit{ \textcolor{red}{Yes it's a great job} } &   \\
\noalign{\smallskip}\svhline\noalign{\smallskip}
\end{tabular}

$^*$ Shows example of ASR error, "lis" is mistranslated into "dis" and it completely changes the sentence meaning and thus conversation flow.

In the first conversation we observe that the model never tried to answer the user's utterance and seems to generate random outputs. In the second conversation, we have a case of repetitiveness and dullness as the bot gets boring insisting on what is probably its personality traits. In the last conversation, the bot quickly stops following user's conversation and starts to be contradictory.
\end{table}

\subsubsection{Impact of Additional Modules}

In order to integrate the effect of orality, an additional, ready to use \texttt{Google API} module for ASR  was involved. This module depending on the recording condition can produce erroneous transcriptions of the users inputs with a direct impact on the conversation flow as it often misleads the text-based model.

Regarding the usage of NMT, the translated dataset for $L_T$ models seemed to have extremely impacted the monolingual PLM \texttt{GPT\_FR} as its English counterpart performs much better. However, for the multilingual model \texttt{BLOOM}, it turns out it had a lower effect as the $L_T$ \textit{TrainOnTarget} model outperforms the others including its $L_S$ counterparts. This may be due to the usage of NMT modules at inference with a model trained on high quality data. We believe that the multilinguality helped BLOOM during training on noisy data especially as it has translation abilities.

 \subsubsection{MAD-X Architecture for Dialogue}
The two models implementing this architecture are the worst performing in this experiment. Although having strongly improved on automatics metrics compared to state-of-the-art models with similar approach. This finds explanation on the adapters being probably under-trained, especially the language adapters. Nevertheless, this approach is not to dismiss as its displayed glimpse of an ability to perform cross-lingual dialogue. Indeed, we notice that \texttt{madxBLOOM\_fr} if given input in $L_S$ was able to give a response in $L_T$ keeping the context of the user's $L_S$ utterance. We believe that further training of models with this architecture should improve their performance and help build cross-lingual chitchat models. The latter is not further investigated in this study but can be of interest for future work.

\section{Conclusion}
The development of French open-domain dialogue models is still far behind its English or even Chinese counterparts nowadays. So is the case for many other languages. The main reason being the lack of specialized datasets. However, the availability of PLMs in this language and advanced NMT tools are assets that can be leveraged to exploit the state-of-the-art from a higher resource language for this task. In this line we evaluated three different approaches and compared the models obtained and also to an English reference model. The TrainOnTarget strategy with a multilingual model, here BLOOM, yielded the best results on human evaluation. It opens the way for future work on more automatically translated data with multilingual models like BLOOM  which implicitly possesses translation abilities. Improvement on the learning objectives then may help catch the reference models in high resource language for this task despite the obstacle of language specific dataset scarcity. Finally, the fact that, outside of chitchat dialog, French is high-resource is not totally limiting on these approaches. Indeed our best model was based on the TrainOnTarget approach with a multilingual model in particular BLOOM which includes some of the lowest-resources languages and is an open-access resource.

\begin{acknowledgement}

We thank the coworkers in the Laboratoire Informatique d'Avignon (LIA) in Avignon University who volunteered during the evaluation process of our models.  The work in this study is supported by $\mu$DialBot project funded by the French National Research Agency (\textit{Agence Nationale de Recherche, ANR}) under the grant \texttt{ANR-20-CE33-0008}. 

\end{acknowledgement}

\bibliographystyle{spmpsci}
\bibliography{iwsds23}
\newpage
\section*{Appendix} 
\addcontentsline{toc}{section}{Appendix}

Rather than performing self-chat pairwise human comparison like in~\cite{Zhaojiang2020} we  chose to involve human during dialogues generation and annotation. In addition, we believe  pairwise evaluation exclude the eventuality to conclude that the two compared models are poor as evaluators have to chose one over the other.
\runinhead{Dialogues collection}
During the first phase, three models were deployed using a modified version of the RASA-X~\cite{DBLP:journals/corr/abs-1712-05181} platform: \texttt{GPT-fr},   \texttt{TransferTransfo (GPT)} and \texttt{BlenderBot 1} - the reference model in this experiment. The two previous in $L_S$ were combined with a \texttt{Google Translate} NMT module at input and output. 4 volunteers were asked to each do a minimum of 15 conversations (to have at least 5 conversations per model per person)  without knowing which model they were interacting with. To avoid redundant conversations, testers were asked to start conversation with a different sentence. Also some personas from the PersonaChat dataset were randomly assigned to the testers in the case the wanted some inspiration to start a conversation. Each conversation should last at least 20 back-and-forths unless the model starts to hallucinate\footnote{A definition of hallucination was provided to the volunteers in the user guide}~\cite{Ji2022}; in this case it was asked to add 2 more inputs before closing the conversation if it didn't get better. This resulted in 60 conversations.

In the second phase, the exact same guidelines were given and this time we deployed the four models based on \texttt{BLOOM}. Again with 4 testers (different from the previous), we collected another 80 conversations. We gathered a total of 140 conversations generated by $L_T$ native speakers in $L_T$.

\runinhead{Dialogues annotations}
Dialog-level evaluation was performed in order to evaluate each conversation individually and completely. Each conversation was evaluated by three annotators from a different batch of volunteers (10), rating from 1 to 5 each of the following criteria based on~\cite{Mehri2020, Ji2022-kr, roller2020recipes}:
\begin{itemize}
\item \textbf{Coherence:} are there hallucinations?, the quality of the bot's expression, coherent answer even if not factual, is the personality the same from the beginning to the end of the dialogue?, does it tend to change the subject too often?
\item \textbf{Engagingness:} does the bot tend to be engaged in the conversation?, does it give constructive and not too vague answers (``okay",``yes",``maybe", ``?" etc.)?, is it willing to restart the conversation when it stalls?
\item\textbf{Humanness:} how much does it feel like a human to human discussion?, is the system repetitive?
\end{itemize}

\runinhead{Inter-annotator agreement:} We can see in Table \ref{tab:A.10} that we have a fair to moderate agreement overall in each category with engagingness the lowest. However the trends seem to be slightly different among given model's conversations, with for instance a low agreement in coherence for the reference model BlenderBot. This  means that evaluators struggles to agree on what a good coherence is worth in rating which is the opposite for the madxBLOOM models established  as the worst were there is a stronger agreement across the abilities.

\begin{table}
\caption{Fleiss-$\kappa$ ~\cite{Falotico2015FleissKS} per ability for each model and overall}
\label{tab:A.10}   
\begin{tabular}{p{3cm}p{3cm}p{3cm}p{2cm}}

\noalign{\smallskip}\svhline\noalign{\smallskip}
Model & Coherence-$\kappa$ & Engagingness-$\kappa$ & Humannes-$\kappa$  \\
\noalign{\smallskip}\svhline\noalign{\smallskip}
BlenderBot & 0.187 & 0.313 & 0.417 \\
\hline\noalign{\smallskip}
GPT\_FR & \textbf{0.521} & 0.229 & 0.375 \\
GPT\_EN  & 0.292  & 0.292 & \textbf{0.521} \\
madxBLOOM\_fr & \textbf{0.423}  & \textbf{0.487} & \textbf{0.423} \\
madxBLOOM\_en & \textbf{0.405}  & 0.286 & \textbf{0.524} \\
BLOOM\_fr  & 0.219  & 0.271  & 0.193 \\
BLOOM\_en  &  \textbf{0.528}  & 0.278  & 0.167 \\
Overall   &  0.361  & 0.301  & 0.379 \\
\noalign{\smallskip}\hline\noalign{\smallskip}
\end{tabular}
\end{table}

\end{document}